\documentclass{article}
\usepackage{spconf,amsmath,graphicx}
\usepackage{booktabs} 
\usepackage{color}
\usepackage[svgnames]{xcolor}
\usepackage{multirow} 
\usepackage{colortbl}  
\usepackage[colorlinks, linkcolor=red, anchorcolor=blue, citecolor=green]{hyperref}

\def\eg{\emph{e.g}.}

\title{Real-time Monocular Depth Estimation on Embedded Systems}
\name{
    Cheng~Feng$^1$,
    Congxuan~Zhang$^{2*}$,
    Zhen~Chen$^{1,2*}$,
    Weiming~Hu$^3$,
    and Liyue~Ge$^1$
    \thanks{
        $^*$Corresponding authors
        }
    }
\address{
    $^1$School of Instrumentation and Optoelectronic Engineering, Beihang University, China\\
    $^2$School of Measuring and Optical Engineering, Nanchang Hangkong University, China\\
    $^3$NLPR, Institute of Automation, Chinese Academy of Sciences, China
    }

\begin{document}
%
\maketitle
\begin{abstract}
    Depth sensing is of paramount importance for unmanned aerial and autonomous vehicles. Nonetheless, contemporary monocular depth estimation methods employing complex deep neural networks within Convolutional Neural Networks are inadequately expedient for real-time inference on embedded platforms. This paper endeavors to surmount this challenge by proposing two efficient and lightweight architectures, RT-MonoDepth and RT-MonoDepth-S, thereby mitigating computational complexity and latency. Our methodologies not only attain accuracy comparable to prior depth estimation methods but also yield faster inference speeds. Specifically, RT-MonoDepth and RT-MonoDepth-S achieve frame rates of 18.4\&30.5 FPS on NVIDIA Jetson Nano and 253.0\&364.1 FPS on Jetson AGX Orin, utilizing a single RGB image of resolution 640$\times$192. The experimental results underscore the superior accuracy and faster inference speed of our methods in comparison to existing fast monocular depth estimation methodologies on the KITTI dataset.
\end{abstract}
%

\section{Introduction}
\label{sec:intro}

    Depth estimation is a crucial perception component for autonomous systems and can also support high-level vision tasks. For example, in the case of robot navigation, obstacle detection, or visual odometer often rely on depth information. While sensors such as LiDAR, stereo cameras, and millimeter-wave radar can provide accurate but sparse depth information, they can be expensive and bulky. In contrast, monocular depth estimation is a low-cost and easy-to-deploy approach to autonomous and intelligent systems that has recently advanced due to the progress in deep learning and reinforcement learning.

\begin{figure}[t]
    \centering
    \centerline{\includegraphics[width=0.95\linewidth]{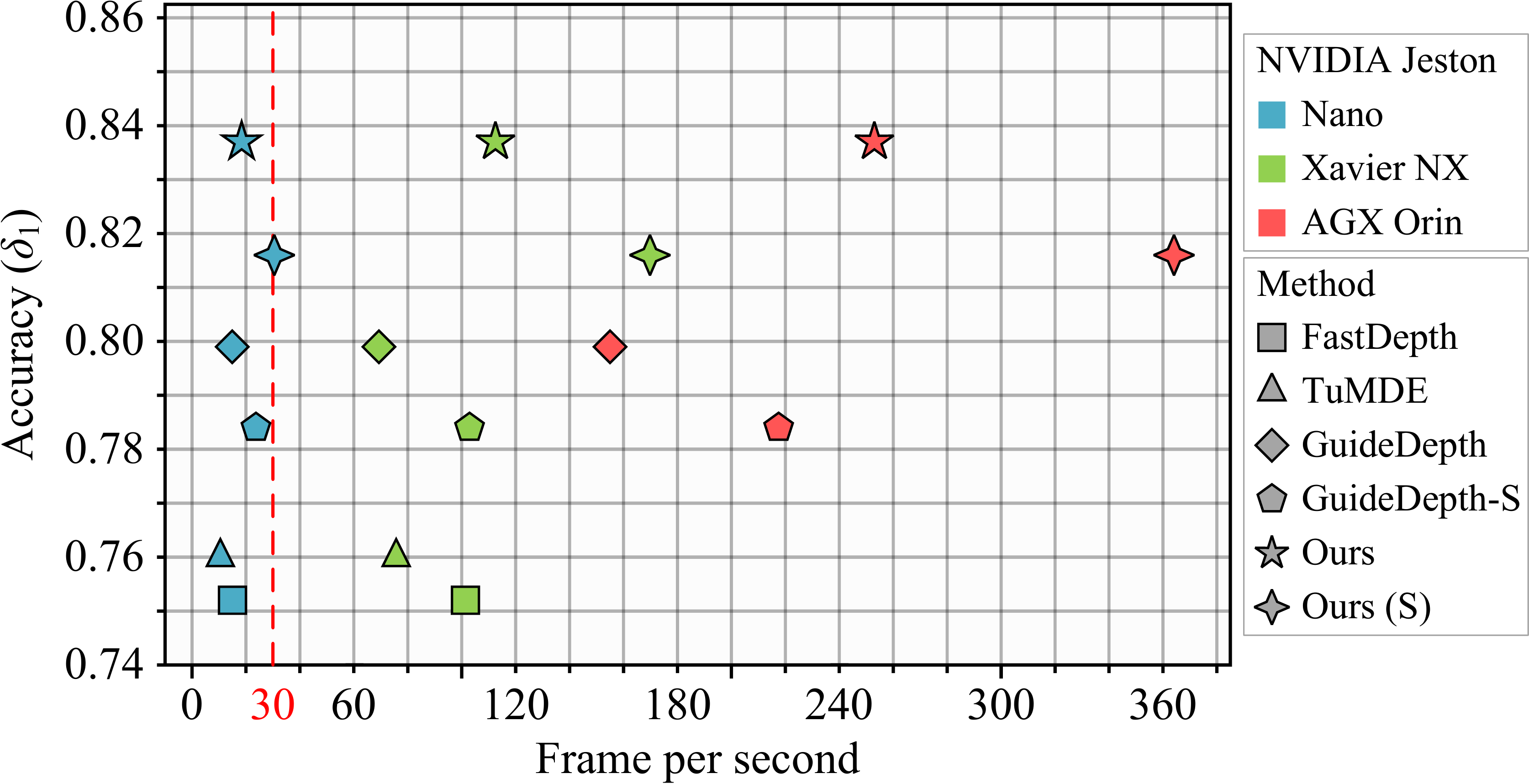}}
    \caption{Accuracy (in $\delta_1$) \textit{vs.} runtime (in FPS) on NVIDIA Jetson Nano, Xavier NX, and AGX Orin for various depth estimation algorithms on KITTI~\cite{DBLP:journals/ijrr/GeigerLSU13} dataset using the Eigen split~\cite{DBLP:conf/nips/EigenPF14}. The top right represents the desired characteristics of a depth estimation network design: high inference speed and high accuracy. The data comes from published papers or is measured using official implementation (online test, $batchsize=1$).}
    \label{fig:1}
\end{figure}
    Considerable monocular depth estimation methods are proposed based on convolutional neural networks and different types of supervision. However, current state-of-the-art CNN-based monocular depth estimation methods~\cite{DBLP:conf/iccv/Yang0DS021,DBLP:conf/cvpr/MiangolehDMPA21} have primarily focused on achieving high accuracy, resulting in computationally intensive methods. Although these methods have significantly improved accuracy, achieving it comes at the cost of increased computational complexity, making them impractical for resource-constrained embedded devices.
    Since most embedded systems have limited computing resources and run multiple concurrent subtasks, achieving real-time performance with a limited computing budget while balancing latency and accuracy is a key challenge for practical deployment of monocular depth estimation in autonomous embedded systems.
    
    Most recent works on efficient depth estimation approaches are devoted to improving the real-time performance of existing monocular depth estimation methods, focusing on the hardware-specific compilation, quantization, and model compression. In addition, some works~\cite{DBLP:conf/icra/WofkMYKS19,DBLP:journals/tii/TuXLLXHY21} use the existing lightweight backbone design to achieve faster execution, while others~\cite{DBLP:conf/icra/0006DGNB22} design an efficient decoding structure to reduce latency.

    This paper introduces a monocular depth estimation method suitable for embedded platforms that aims to balance depth estimation accuracy and latency for autonomous systems. Two efficient encoder-decoder network architectures, RT-MonoDepth and RT-MonoDepth-S, are proposed, which focus on high-accuracy with medium-latency design and medium-accuracy with low-latency design, respectively. The middle and low latency network designs, RT-MonoDepth and RT-MonoDepth-S, can perform real-time depth estimation on NVIDIA Jetson Nano\footnote{https://developer.nvidia.com/embedded/jetson-nano}, Xavier NX\footnote{https://developer.nvidia.com/embedded/jetson-xavier-nx}, and AGX Orin\footnote{https://www.nvidia.com/en-us/autonomous-machines/embedded-systems/jetson-orin/} without post-processing. Fig.~\ref{fig:1} illustrates that RT-MonoDepth can achieve reasonable accuracy and operate at over 253.0 frames per second (FPS) on AGX Orin and around 18.4 FPS on Nano. Moreover, RT-MonoDepth-S delivers solid performance at a high frame rate, operating at over 364.1 FPS on AGX Orin and around 30.5 FPS on Nano.

    In summary, we propose two lightweight convolutional neural networks for monocular depth estimation. To the best of the authors’ knowledge, our approach outperforms the related methods that target fast monocular depth estimation in terms of accuracy for the KITTI~\cite{DBLP:journals/ijrr/GeigerLSU13} benchmark while achieving the fastest inference speed.

\section{Related Work}
    Depth sensing from a monocular camera has been an active topic in the research of computer vision and robotics communities for over a decade. 
    Eigen~et~al.~\cite{DBLP:conf/nips/EigenPF14} introduced the first CNN-based monocular depth estimation network, which utilizes a coarse-to-fine network with two stacked sub-networks. One sub-network predicts the global coarse depth map, and the other refines local details. 
    In contrast, Laina~et~al.~\cite{DBLP:conf/3dim/LainaRBTN16} suggested an end-to-end encoder-decoder architecture for monocular depth estimation, using a pre-trained ResNet~\cite{DBLP:conf/cvpr/HeZRS16} backbone as a feature extractor, and got comparable results to the depth sensor.

    Godard~et~al.~\cite{DBLP:conf/cvpr/GodardAB17} proposed the first unsupervised training strategy, which utilizes the left-right consistency of stereo pairs to address the problem of CNN model training requiring a large amount of labeled data. 
    Meanwhile, Zhou~et~al.~\cite{DBLP:conf/cvpr/ZhouBSL17} introduced an unsupervised framework that solves the depth estimation problem as a view synthesis problem by combining the depth model with a pose model. 
    Furthermore, Godard~et~al.~\cite{DBLP:conf/iccv/GodardAFB19} suggested a self-supervised strategy by applying auto-masking to moving objects and using novel minimum reprojection loss to improve accuracy further, making Monodepth2 a significant baseline model.
    
    Most recently, some studies aim to find a trade-off between accuracy and real-time performance. Initially, Wofk~et~al.~\cite{DBLP:conf/icra/WofkMYKS19} proposed FastDepth, an efficient method composed of a pre-trained MobileNetV2~\cite{DBLP:conf/cvpr/SandlerHZZC18} backbone with a lightweight decoder. 
    To increase inference speed, Wofk~et~al.~\cite{DBLP:conf/icra/WofkMYKS19} and Tu~et~al.~\cite{DBLP:journals/tii/TuXLLXHY21} compiled their models with TVM~\cite{DBLP:conf/osdi/ChenMJZYSCWHCGK18}, and employed refine network or reinforcement learning algorithm to reduce model size, respectively. 
    In the latest research, Rudolph~et~al.~\cite{DBLP:conf/icra/0006DGNB22} propose an efficient network, GuideDepth, composed of a DDRNet-23-slim~\cite{DBLP:journals/corr/abs-2101-06085} backbone and a lightweight decoder with a guided upsampling block. GuideDepth is compared with Wofk~et~al.~\cite{DBLP:conf/icra/WofkMYKS19} and Tu~et~al.~\cite{DBLP:journals/tii/TuXLLXHY21} in the condition without post-processing and achieves the best result in both accuracy and inference speed on embedded platforms.

\begin{figure*}[!t]
    \centering
    \includegraphics[width=0.95\textwidth]{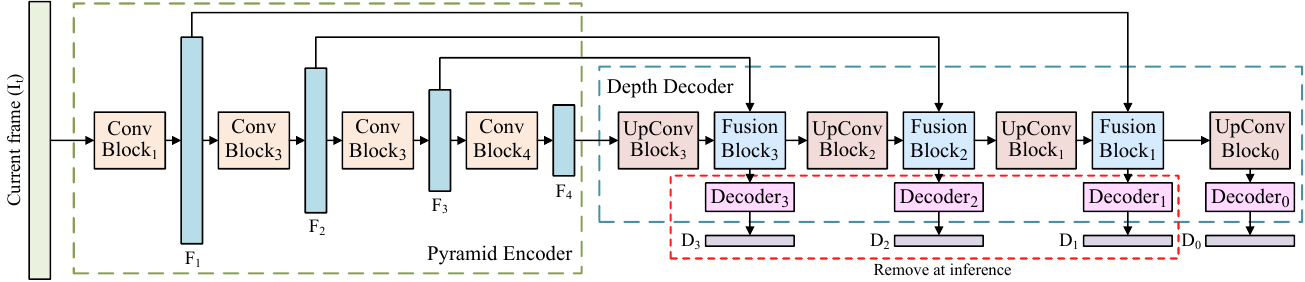}
    \caption{Proposed RT-MonoDepth framework. The shape of input image is $H \times W \times 3$, symbol 'F' denotes feature maps in the pyramid encoder, symbol 'D' denotes predicted depth maps in each scale, the subscript number indicates their shape (\eg, $F_n\colon\frac{H}{2^n} \times \frac{W}{2^n} \times C_n$, $D_n\colon\frac{H}{2^n} \times \frac{W}{2^n} \times 1$). The Decoder$_3$, Decoder$_2$ and Decoder$_1$ can be removed when inferring.}
    \label{fig:2}
\end{figure*}
\section{Methodology}
    Our proposed RT-MonoDepth framework is shown in Fig.~\ref{fig:2}. It consists of two parts: a pyramid encoder and a depth decoder. The pyramid encoder extracts high-level low-resolution features from the input image. These feature maps are then transmitted into the depth decoder, where they are gradually upsampled, fused, and used to predict the depth map at each scale. We pursue low latency and high-precision designs for both the pyramid encoder and the depth decoder in order to design a monocular depth estimation network that can run in real-time on embedded systems.
    
    \subsection{Lightweight Pyramid Encoder}
    The pyramid encoder in past monocular depth estimation methods usually employed a designed backbone network for image classification. ResNet~\cite{DBLP:conf/cvpr/HeZRS16} and MobileNet~\cite{DBLP:conf/cvpr/SandlerHZZC18} were commonly used due to their powerful feature extraction ability and high accuracy. However, these structures are subject to limitations of complexity and latency, making them difficult to deploy to resource-constrained embedded devices. Moreover, it is hard to design a flexible decoder for low latency due to the fixed shape of the backbone network.
    
    ResNet and MobileNet insert the batch normalization layer to accelerate convergence due to their complex topologies and deep layers. However, a small batch size brings instability to the model using batch normalization~\cite{DBLP:conf/eccv/WuH18}, and the capacity of GPU memory limits the dense regression task. Since we aim to design a simple, lightweight encoder, removing the normalization layer has no negative impact and can reduce computation slightly.

    MobileNet reduces the computational complexity in theory by using depth-wise convolution instead of standard convolution. However, current implementations of deep learning frameworks cannot fully utilize the GPU without hardware support, making depth-wise convolution slow during both training and inference~\cite{DBLP:conf/ijcnn/QinZLZP18}.

    Taking the above two points into consideration, we construct a 4-layer pyramid encoder from scratch that targets low latency. Each layer features a \textit{ConvBlock} which downsamples inputs and produces higher-level feature maps. The \textit{ConvBlock} consists of three $3 \times 3$ standard convolution layers without skip-connections or batch normalization layers, in order to minimize latency.

    \subsection{Simplified Depth Decoder}
    The function of the decoder is to upsample and fuse the output of the encoder to form a dense depth prediction. Our main objective is to determine how to design the three most critical parts of the depth decoder - upsampling layers, fusion layers, and prediction layers - in order to improve throughput and decrease the latency.
    
        \textit{Upsampling:} Upsampling operation is a fundamental design element of the low-latency depth decoder. Over the past few decades, various upsampling methods have been proposed, such as unpooling, transpose convolution, interpolation, and interpolation combined with convolution. FastDepth~\cite{DBLP:conf/icra/WofkMYKS19} explored the impact of different upsampling methods on depth performance. They compared different combinations of upsampling methods and ultimately proposed an upsampling method called NNConv5, which consists of $5 \times 5$ depth-wise separable convolution followed by nearest-neighbor interpolation with a scale factor of 2. To achieve an even faster inference speed, we have substituted NNConv5 with NNConv3. Each NNConv3 \textit{UpConvBlock} performs $3 \times 3$ standard convolution and reduces the number of output channels by $\frac{1}{2}$ relative to the number of input channels, and followed a nearest-neighbor interpolation that doubles the resolution of intermediate feature maps.
\begin{figure*}[!t]
    \centering
    \includegraphics[width=0.8\textwidth]{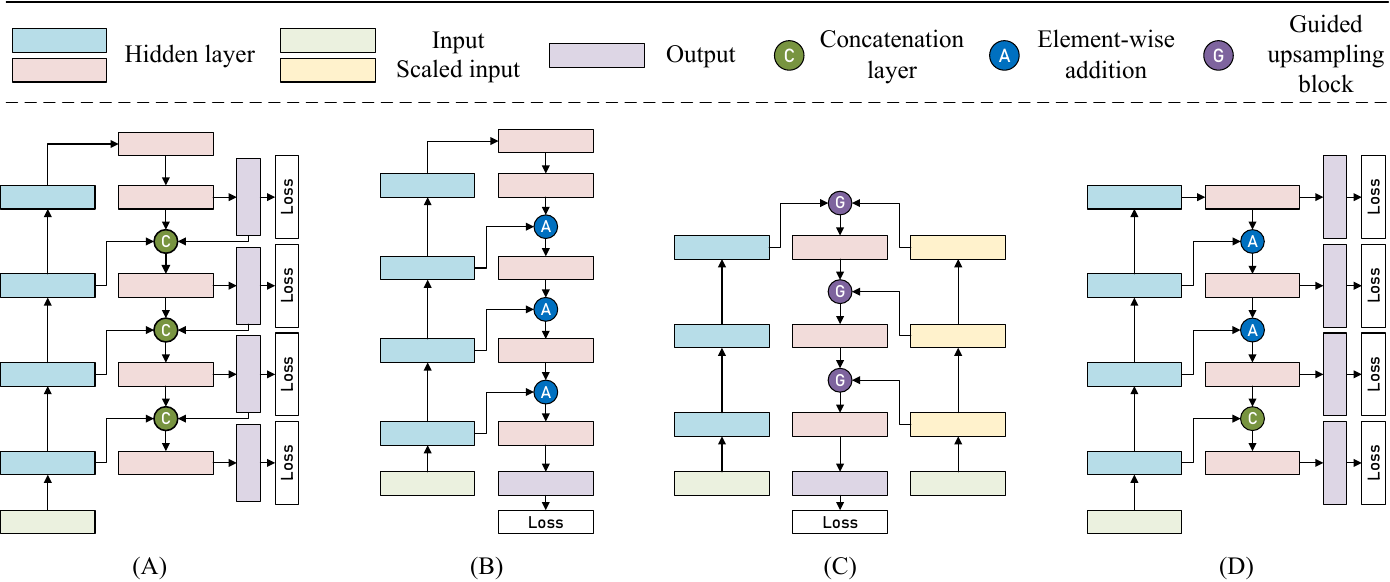}
    \caption{Feature fusion methods of different approaches. (A) Monodepth2~\cite{DBLP:conf/iccv/GodardAFB19}; (B) FastDepth~\cite{DBLP:conf/icra/WofkMYKS19}; (C) GuideDepth~\cite{DBLP:conf/icra/0006DGNB22}; (D) Our model.
    }
    \label{fig:3}
\end{figure*}
        
        \textit{Fusion:} Fusing features of different scales is an essential means to improve performance in many works~\cite{DBLP:conf/eccv/ZhangZPXS18}. Low-level features with high resolution contain more spatial information but less semantic information and more noise. On the other hand, high-level features with low resolution contain more semantic information but less spatial information. Two standard fusion methods in deep learning practice are element-wise addition and concatenation. Monodepth2~\cite{DBLP:conf/iccv/GodardAFB19} gradually concatenates upper-level feature maps to lower levels, while FastDepth~\cite{DBLP:conf/icra/WofkMYKS19} gradually adds upper-level feature maps to the next level. In contrast, GuideDepth~\cite{DBLP:conf/icra/0006DGNB22} introduces the guided upsampling block, which uses the input image as guidance to upsample feature maps without previous level feature maps. To balance latency and accuracy, we perform element-wise addition at the first and second layers to accelerate feature fusion, and using concatenate operation at the third layer of the depth decoder for better final accuracy.

        \textit{Prediction:} Our model sets one prediction decoder at each scale. Each decoder consists of two $3 \times 3$ convolution layers, followed by a leakyReLU and a sigmoid function as activating functions, respectively. The decoders receive the fused feature maps and yield depth predictions at each scale for self-supervision. To minimize latency, we do not pass the intermediate depth prediction to the next level. This allows us to disable the first three decoders to speed up inference.

\section{Experiments}
    In this section, we present experimental results to demonstrate the effectiveness of our approach. We first introduce the details of our implementation and then evaluate our method against existing high-precision or low-latency methods on the KITTI~\cite{DBLP:journals/ijrr/GeigerLSU13} benchmark for monocular depth estimation. We also provide ablation studies of our framework based on accuracy and latency metrics. Finally, we discuss the limitations of our approach and suggest future directions for improvement.

    \subsection{Implementation details}
        
        \textit{Hardware Platforms:} We train our networks on a single NVIDIA RTX 3090 GPU with 9 GB of GPU memory, and evaluate the real-time performance on three edge computing embedded systems: the NVIDIA Jetson Nano, Xavier NX, and AGX Orin. The Jetson Nano and Xavier NX are low-power embedded devices, while the AGX Orin is the most powerful platform deployed in autonomous systems. We report results for Jetson Nano in 10W power mode and Xavier NX in 15W power mode, using the same configurations as described in~\cite{DBLP:conf/icra/0006DGNB22}. Additionally, we present detailed results for the AGX Orin to highlight our advantage compared to state-of-the-art fast depth estimation methods.
        
        \textit{Dataset:} We train and evaluate our networks on the KITTI~\cite{DBLP:journals/ijrr/GeigerLSU13} dataset, following the train/test data split described in~\cite{DBLP:conf/nips/EigenPF14}. , We remove static frames~\cite{DBLP:conf/cvpr/ZhouBSL17} during training, and clip the depth to 80m per standard practice~\cite{DBLP:conf/cvpr/GodardAB17} during evaluating.
        
        \textit{Training Details:} Our method is implemented in PyTorch using 32-bit floating-point precision. We use the AdamW optimizer with a learning rate of $1 \times 10^{-4}$ and a batch size of 8. We train for 20 epochs, reducing the learning rate by 10 after 15 epochs. Data augmentation and self-supervised strategy during training are the same as in Monodepth2~\cite{DBLP:conf/iccv/GodardAFB19}.
        
        \textit{Testing Details:} There are two standard testing methods in deep learning: offline and online. Offline testing is used when real-time requirements are not necessary, allowing multiple data ($batchsize=N$) to be processed in parallel. In contrast, online testing only allows one data ($batchsize=1$) to be processed at a time to test real-time performance. For a fair comparison, all inference speeds in our experiments are reported on models converted to 16-bit floating-point precision TensorRT\footnote{https://developer.nvidia.com/tensorrt} with online testing. Unlike~\cite{DBLP:conf/icra/0006DGNB22}, we warm up each method by running it 1000 times and then measure the time consumption with an average test time of 5000 iterations (10 warm-up and 200 testing in~\cite{DBLP:conf/icra/0006DGNB22}).
        
        \textit{Evaluation Metrics:}
            We use standard six metrics used in 
            used in prior work~\cite{DBLP:conf/iccv/GodardAFB19} to compare various methods. Let $\hat{d_i}$ be the pixel of prediction depth map $\hat{d}$, and $d_i$ be the pixel in the ground-truth $d$, and $N$ be the total number of pixels. Then we use the following evaluation metrics: 
            absolute relative error ($\mathrm{AbsRel}$): 
            $\frac{1}{N}\sum\frac{| d_i-\hat{d_i} |}{d_i}$;
            square relative error ($\mathrm{SqRel}$):
            $\frac{1}{N}\sum\frac{| d_i-\hat{d_i} |^2}{d_i}$;
            root mean square error ($\mathrm{RMSE}$):
            $\sqrt{\frac{1}{N}\sum{| d_i-\hat{d_i} |}^2}$;
            root mean square logarithmic error ($\mathrm{RMSE_{log}}$):
            $\sqrt{\frac{1}{N} \sum|\log _{10} d_{i}-\log _{10} \hat{d}_{i}|^{2}}$;
            threshold accuracy ($\delta_j$):
            $\%$ of $d_i$ s.t. $\max(\frac{d_i}{\hat{d_i}},\frac{\hat{d_i}}{d_i})=\delta<thr$ for $thr=1.25,1.25^2,1.25^3$.
\begin{table*}
\caption{Quantitative results of depth estimation on the KITTI dataset (Eigen split, 640$\times$192, 0-80m).}
\label{table:Quantitative_results}
\renewcommand{\arraystretch}{1.05}
\setlength\arrayrulewidth{1pt} 
\definecolor{COLOR1}{RGB}{232, 160, 147}
\definecolor{COLOR2}{RGB}{147, 186, 232}
\centering
\resizebox{\linewidth}{!}{
\begin{tabular}{|c|c|c|c||c|c|c|c|c|c|c|c|c|c|}
\hline
methods   & Data & \begin{tabular}[c]{@{}c@{}}pre-\\train\end{tabular} & \begin{tabular}[c]{@{}c@{}}\#para\\~(M)\end{tabular} & $\cellcolor{COLOR1}\mathrm{AbsRel}\downarrow$ & $\cellcolor{COLOR1}\mathrm{SqRel}\downarrow$ & $\cellcolor{COLOR1}\mathrm{RMSE}\downarrow$ & $\cellcolor{COLOR1}\mathrm{RMSE_{log}}\downarrow$ & $\cellcolor{COLOR2}\delta<1.25\uparrow$ & $\cellcolor{COLOR2}\delta<1.25^2\uparrow$ & $\cellcolor{COLOR2}\delta<1.25^3\uparrow$  & \cellcolor{COLOR2}\begin{tabular}[c]{@{}c@{}}FPS$\uparrow$\\~(Nano)\end{tabular} & \cellcolor{COLOR2}\begin{tabular}[c]{@{}c@{}}FPS$\uparrow$\\~(Xavier)\end{tabular} & \cellcolor{COLOR2}\begin{tabular}[c]{@{}c@{}}FPS$\uparrow$\\~(Orin)\end{tabular}  \\
\hline
\hline
FastDepth~\cite{DBLP:conf/icra/WofkMYKS19}     & D   & $\bullet$  & 4.0  & 0.168             & -                 & 5.839             & -               & 0.752             & 0.927             & 0.977             & \underline{15.0} & \underline{101.4}  & - \\
TuMDE~\cite{DBLP:journals/tii/TuXLLXHY21}      & D   & $\bullet$  & 5.7  & \underline{0.150} & -                 & 5.801             & -               & 0.760             & 0.930             & 0.980             & 10.5             & 75.7               & - \\
GuideDepth~\cite{DBLP:conf/icra/0006DGNB22}    & D   & $\bullet$  & 5.8  & \textbf{0.142}    & -                 & \textbf{5.194}    & -               & \textbf{0.799}    & \textbf{0.941}    & \textbf{0.982}    & 15.2             & 69.3               & 155.0 \\
GuideDepth-S~\cite{DBLP:conf/icra/0006DGNB22}  & D   & $\bullet$  & 5.7  & \textbf{0.142}    & -                 & \underline{5.480} & -               & \underline{0.784} & \underline{0.936} & \underline{0.981} & \textbf{25.2}    & \textbf{102.9}     & 217.5 \\
\hline
\hline
Monodepth2~\cite{DBLP:conf/iccv/GodardAFB19}   & M    &            & 14.3 & 0.132             & 1.044             & 5.142             & 0.210             & 0.845             & 0.948             & 0.977              & -                & -                 & 142.3 \\
Monodepth2~\cite{DBLP:conf/iccv/GodardAFB19}   & M    & $\bullet$  & 14.3 & \textbf{0.115}    & \textbf{0.903}    & \textbf{4.863}    & \textbf{0.193}    & \textbf{0.877}    & \textbf{0.959}    & \textbf{0.981}     & -                & -                 & 142.3 \\
\textbf{RT-MonoDepth}                             & M    &            & 2.8  & \underline{0.125} & \underline{0.959} & \underline{4.985} & \underline{0.202} & \underline{0.857} & \underline{0.952} & \underline{0.979}  & \underline{18.4} & \underline{112.5} & \underline{253.0} \\
\textbf{RT-MonoDepth-S}                           & M    &            & 1.2  & 0.132             & 0.997             & 5.262             & 0.214             & 0.840             & 0.946             & 0.977              & \textbf{30.5}    & \textbf{169.8}    & \textbf{364.1} \\
\hline
\hline
Monodepth2~\cite{DBLP:conf/iccv/GodardAFB19}   & MS   &            & 14.3  & \underline{0.127} & 1.031             & 5.266             & 0.221             & 0.836             & 0.943             & 0.974             & -                & -                 & 142.3 \\
Monodepth2~\cite{DBLP:conf/iccv/GodardAFB19}   & MS   & $\bullet$  & 14.3  & \textbf{0.106}    & \textbf{0.818}    & \textbf{4.750}    & \textbf{0.196}    & \textbf{0.874}    & \textbf{0.957}    & \textbf{0.979}    & -                & -                 & 142.3 \\
\textbf{RT-MonoDepth}                             & MS   &            & 2.8   & \underline{0.127} & \underline{0.995} & \underline{5.153} & \underline{0.217} & \underline{0.837} & \underline{0.944} & \underline{0.975} & \underline{18.4} & \underline{112.5} & \underline{253.0} \\
\textbf{RT-MonoDepth-S}                           & MS   &            & 1.2   & 0.135             & 1.040             & 5.413             & 0.229             & 0.816             & 0.936             & 0.973             & \textbf{30.5}    & \textbf{169.8}    & \textbf{364.1} \\
\hline
\end{tabular}
}
\end{table*}
    \subsection{Final Results and Comparison With Prior Work}
        Table~\ref{table:Quantitative_results} summarizes the results of our methodology compared to existing high-precision or low-latency methods. The 'D' symbol in the Data column indicates methods that use depth supervision during training, 'M' denotes self-supervised training on monocular video, and 'MS' represents models trained with stereo video. We also include results for Monodepth2 with ImageNet pretraining, indicated by a checkmark in the pre-train column.

        We compare our model to low-latency methods such as FastDepth~\cite{DBLP:conf/icra/WofkMYKS19}, TuMDE~\cite{DBLP:journals/tii/TuXLLXHY21}, and GuideDepth~\cite{DBLP:conf/icra/0006DGNB22}, as well as the classical method Monodepth2~\cite{DBLP:conf/iccv/GodardAFB19}. We also evaluate a smaller model called RT-MonoDepth-S, which drops the \textit{FusionBlock} and reduces the layers of \textit{ConvBlock} in the original model, resulting in about 6\% less accuracy but 47\% faster inference time than RT-MonoDepth.

        As shown in Table~\ref{table:Quantitative_results}, RT-MonoDepth achieves an average of 5\% better performance with 53\% fewer parameters and an average of 60\% higher FPS. The RT-MonoDepth-S lite version performs an average of 3\% better than GuideDepth with 75\% fewer parameters and an average of 63\% higher FPS. Compared to the classical method Monodepth2 without pre-training, RT-MonoDepth slightly outperforms Monodepth2 in all metrics. However, there is still a significant performance gap between our method and pre-trained Monodepth2.

        As a result of our efficient and lightweight architecture design, RT-MonoDepth can achieve up to 18.4 FPS on the Jetson Nano and 253.0 FPS on the AGX Orin. RT-MonoDepth-S delivers up to 30.5 FPS on the NVIDIA Jetson Nano and up to 364.1 FPS on the AGX Orin. The low latency makes our methods suitable for deployment on autonomous robot platforms with high real-time performance requirements. As far as we know, our method is the fastest non-pruned monocular depth estimation method on embedded systems.

\begin{figure*}[!t]
    \centering
    \includegraphics[width=0.97\textwidth]{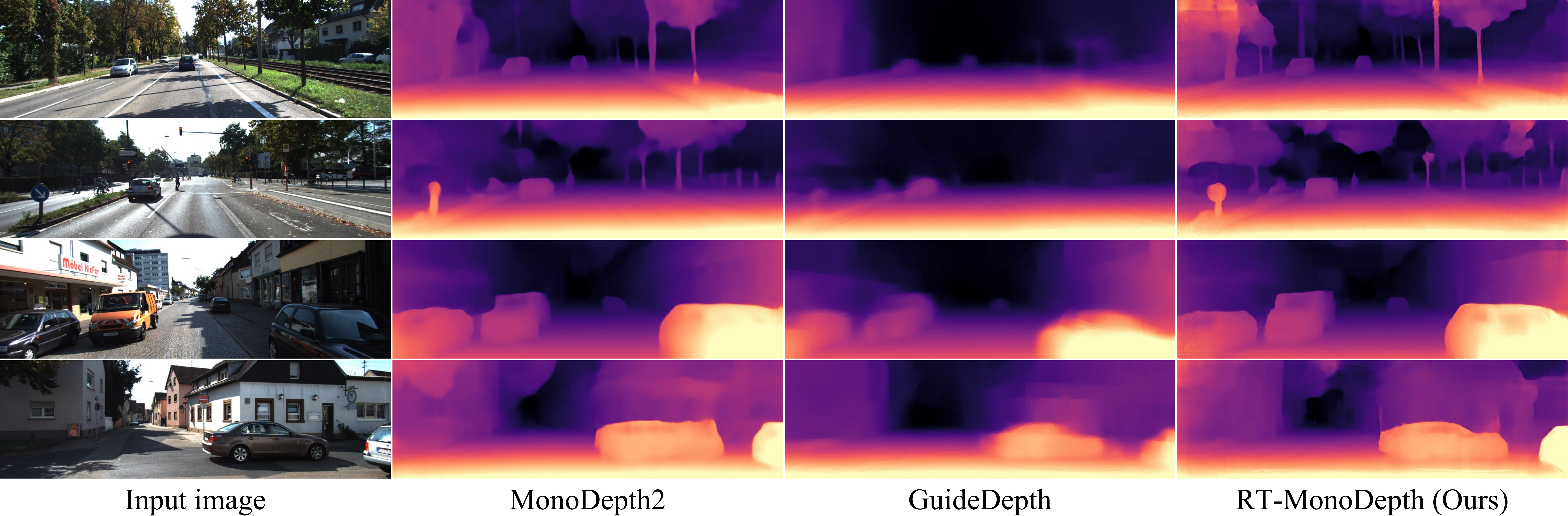}
    \caption{Qualitative results of monocular depth estimation comparing RT-MonoDepth with Monodepth2 and GuideDepth on the KITTI dataset. Our method provides cleaner boundaries and more effective object reconstruction than the other methods.
    }
    \label{fig:4}
\end{figure*}

        Fig.~\ref{fig:4} presents the qualitative results of our RT-MonoDepth method, the state-of-the-art fast monocular depth estimation method GuideDepth, and the classical method Monodepth2 (without pre-training). As shown in Fig.~\ref{fig:4}, our method outperforms GuideDepth significantly and can provide more effective object reconstruction than MonoDepth2, especially in the edge regions.

    \subsection{Ablation Studies}
        We investigate the design space of the RT-MonoDepth framework and perform ablation studies on supervision scales, pyramid encoder levels and feature fusions. All ablation studies are tested on Jetson Nano at a resolution of $640 \times 192$.

        \textit{Supervision scales:} Monodepth2~\cite{DBLP:conf/iccv/GodardAFB19}, FastDepth~\cite{DBLP:conf/icra/WofkMYKS19}, and GuideDepth~\cite{DBLP:conf/icra/0006DGNB22} gradually merge feature maps in the decoder to output a full-resolution depth map at the last level. Monodepth2~\cite{DBLP:conf/iccv/GodardAFB19} predicts depth maps and applies supervision signals at different scales, and these intermediate depth maps are merged upward and cannot be ignored during inference. FastDepth~\cite{DBLP:conf/icra/WofkMYKS19} and GuideDepth~\cite{DBLP:conf/icra/0006DGNB22} only yield depth maps at the last level and do not apply additional intermediate supervision signals. Table~\ref{table:Ablation_results} summarizes the results of four variants that use 1, 2, 3, and 4 layers of supervision. The results indicate that deeper depth supervision leads to consistently better performance. Unlike Monodepth2, the intermediate predictions of RT-MonoDepth will not affect the final real-time performance as they can be ignored during inference.

        \textit{Feature Pyramid Levels:} The feature pyramid levels determine the minimum resolution at which the decoder starts, which is a critical parameter of coarse-to-fine architectures. To fully utilize the employed backbone network, Monodepth2~\cite{DBLP:conf/iccv/GodardAFB19} and FastDepth~\cite{DBLP:conf/icra/WofkMYKS19} extract the feature map with a minimum resolution of $\frac{1}{32}$. On the other hand, GuideDepth~\cite{DBLP:conf/icra/0006DGNB22} only extracts the feature map with a minimum resolution of $\frac{1}{8}$ to achieve faster inference speed. Table~\ref{table:Ablation_results} summarizes the results of four variants that use 2, 3, 4, and 5 levels of the feature pyramid, which extract the feature map with a minimum resolution of $\frac{1}{4}$, $\frac{1}{8}$, $\frac{1}{16}$, and $\frac{1}{32}$, respectively. The results indicate that more levels of the feature pyramid provide higher precision but worse real-time performance. A pyramid level of 4 strikes a balance between accuracy and latency.

        \textit{Feature Fusions:} We evaluate the accuracy and latency of the model under different feature fusion methods. Table~\ref{table:Ablation_results} summarizes the results of four methods that use different feature fusion strategies. The symbol '+' represents element-wise addition, and 'c' stands for concatenation. The results demonstrate that more concatenation operations provide higher precision but worse real-time performance. The '++c' method strikes a balance between accuracy and latency.
\begin{table}
\caption{Ablation experiments. \textbf{Bold}: Final settings.}
\label{table:Ablation_results}
\renewcommand{\arraystretch}{1.05}
\setlength\arrayrulewidth{1pt} 
\definecolor{COLOR1}{RGB}{232, 160, 147}
\definecolor{COLOR2}{RGB}{147, 186, 232}
\centering
\resizebox{0.9\linewidth}{!}{
\begin{tabular}{|c|c||c|c|c|c|}
\hline
Ablation   & Parameter   & $\cellcolor{COLOR1}\mathrm{AbsRel}\downarrow$ & $\cellcolor{COLOR1}\mathrm{RMSE}\downarrow$ & $\cellcolor{COLOR2}\delta<1.25\uparrow$ & \cellcolor{COLOR2}FPS$\uparrow$  \\
\hline
\hline
\multirow{4}{*}{\begin{tabular}[c]{@{}c@{}}Supervision\\Scales\end{tabular}}    & 1             & 0.128   & 5.227   & 0.834   & 18.4  \\
                                                                                & 2             & 0.128   & 5.157   & 0.834   & 18.4  \\
                                                                                & 3             & 0.128   & 5.158   & 0.836   & 18.4  \\
                                                                                & \textbf{4}    & 0.127   & 5.153   & 0.837   & 18.4  \\
\hline
\multirow{4}{*}{\begin{tabular}[c]{@{}c@{}}Encoder\\Levels\end{tabular}}        & 2             & 0.146   & 5.357   & 0.814   & 24.3  \\
                                                                                & 3             & 0.133   & 5.207   & 0.828   & 20.9  \\
                                                                                & \textbf{4}    & 0.127   & 5.153   & 0.837   & 18.4  \\
                                                                                & 5             & 0.123   & 5.080   & 0.837   & 16.2  \\ 
\hline
\multirow{4}{*}{\begin{tabular}[c]{@{}c@{}}Fusion\\Methods\end{tabular}}        & +++           & 0.132   & 5.257   & 0.834   & 19.3  \\
                                                                                & \textbf{++c}  & 0.127   & 5.153   & 0.837   & 18.4  \\
                                                                                & +cc           & 0.126   & 5.139   & 0.837   & 17.6  \\
                                                                                & ccc           & 0.126   & 5.086   & 0.838   & 15.8  \\ 
\hline
\end{tabular}
}
\end{table}

    \subsection{Comparison of inferencing speed}
\begin{figure}[!t]
    \centering
    \includegraphics[width=0.35\textwidth]{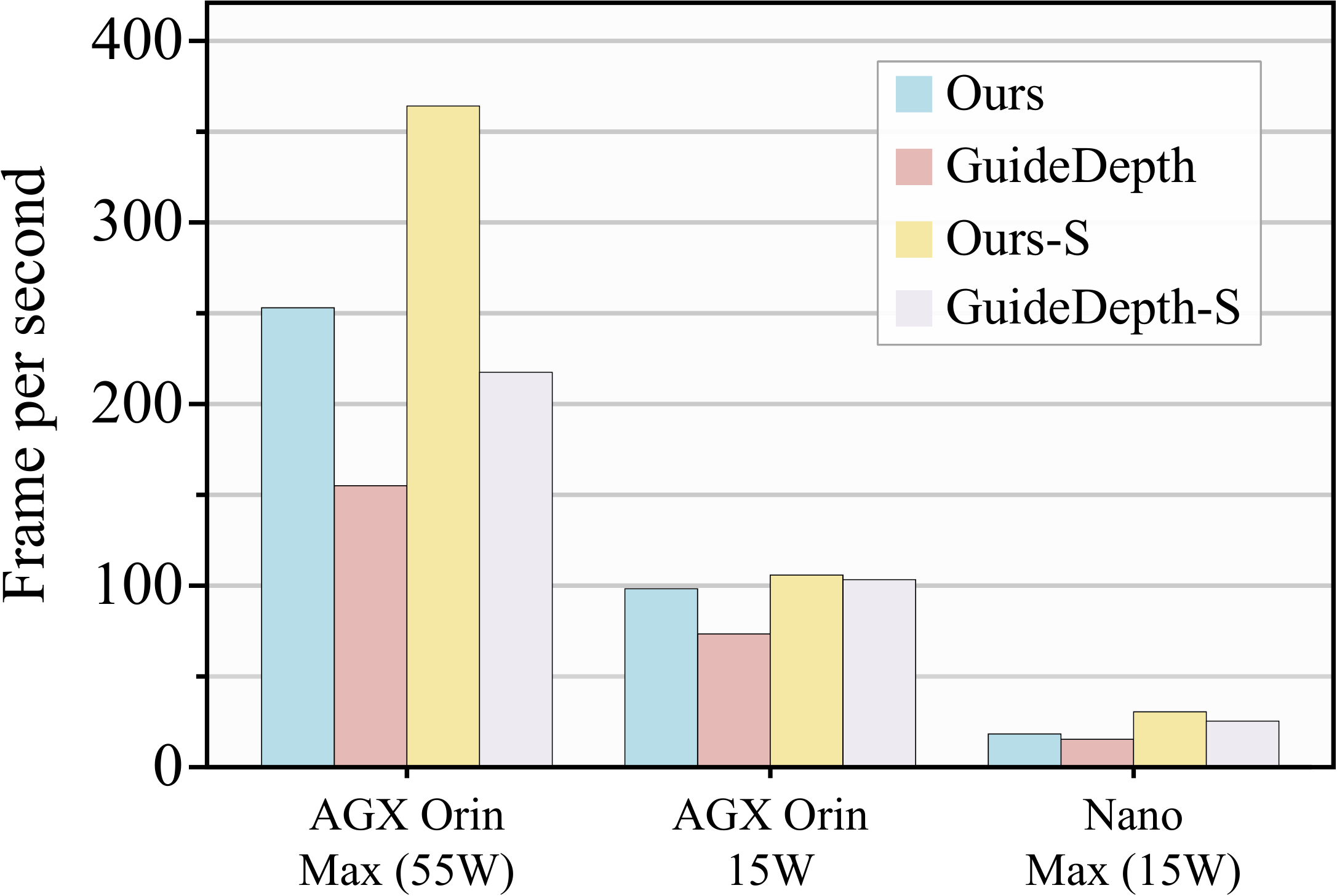}
    \caption{Comparison results of inferencing speed comparing RT-MonoDepth with GuideDepth. Our method provides faster inferencing speed than the GuideDepth.}
    \label{fig:5}
\end{figure}

        We conduct various speed tests of RT-MonoDepth and the state-of-the-art fast monocular depth estimation method GuideDepth on the AGX Orin and Nano using an image resolution of 640$\times$192. The results are presented in Fig.~\ref{fig:5}. In the full power mode (Orin Max and Nano Max), our method outperforms GuideDepth significantly. With GPU optimization, the advantages of our method become more prominent. Furthermore, we set AGX Orin to work in 15W mode to simulate resource-constrained scenarios where a single device runs multiple tasks in parallel. The results demonstrate that our method is faster than GuideDepth, even under conditions of limited computing resources.

\section{LIMITATION}
    Table~\ref{table:Limitation} illustrates the performance of RT-MonoDepth across varying input resolutions. While our method demonstrates promising outcomes at lower resolutions, its efficacy diminishes with high-resolution inputs. Specifically, our model, when trained with high-resolution images, exhibits an approximate 4\% reduction in average performance compared to the low-resolution counterpart. This performance decline aligns with findings in existing research~\cite{DBLP:conf/iccv/Hu0O19,DBLP:conf/cvpr/MiangolehDMPA21}, indicating that monocular depth estimation models heavily rely on learned image edge features. Higher-resolution images necessitate a model with an encoder boasting an expanded receptive field and complexity for optimal learning. Given that our model employs a lightweight, simplistic encoder, accurate learning becomes challenging with high-resolution inputs, contributing to the observed performance degradation.

\begin{table}[!t]
    \caption{Quantitative results of RT-MonoDepth with different resolution images on the KITTI dataset (Eigen split, 0-80m).}
    \label{table:Limitation}
    \renewcommand{\arraystretch}{1.05}
    \setlength\arrayrulewidth{1pt} 
    \definecolor{COLOR1}{RGB}{232, 160, 147}
    \definecolor{COLOR2}{RGB}{147, 186, 232}
    \centering
    \resizebox{0.7\linewidth}{!}{
    \begin{tabular}{|c|c||c|c|c|}
    \hline
    Resolution                                                                   & Data &$\cellcolor{COLOR1}\mathrm{AbsRel}\downarrow$ & $\cellcolor{COLOR1}\mathrm{RMSE}\downarrow$ & $\cellcolor{COLOR2}\delta<1.25\uparrow$  \\
    \hline
    \hline
    \multirow{2}{*}{\begin{tabular}[c]{@{}c@{}}640$\times$192\end{tabular}}      & M    & 0.125              & 4.985            & 0.857               \\
                                                                                 & MS   & 0.126              & 5.157            & 0.836               \\
    \hline
    \multirow{2}{*}{\begin{tabular}[c]{@{}c@{}}1024$\times$320\end{tabular}}     & M    & 0.139              & 5.108            & 0.837               \\
                                                                                 & MS   & 0.136              & 5.182            & 0.825               \\
    \hline
    \end{tabular}}
\end{table}

\section{CONCLUSION}
    In this work, we focus on designing low-latency monocular depth estimation networks for resource-constrained robot systems on embedded platforms. We propose efficient network architectures, RT-MonoDepth and RT-MonoDepth-S, with low complexity and low-latency encoder-decoder design, achieving high frame rates on Jetson Nano, Xavier NX, and AGX Orin while maintaining comparable accuracy with related architectures on the KITTI dataset. Our method is the fastest no-pruning monocular depth estimation method to date. We hope our methods can be practically applied to real-world autonomous driving and robotics. For future work, we aim to improve the accuracy of high-resolution input and further reduce runtime.

\section{ACKNOWLEDGE}
    This work was supported in part by the National Natural Science Foundation of China under Grant 62222206 and Grant 62272209; in part by the Technology Innovation Guidance Program of Jiangxi Province under Grant 20212AEI91005; in part by the Key Research and Development Program of Jiangxi Province under Grant 20232BBE50006; in part by the Major Research and Development Project of Jiangxi Province 20232ACC01007; and in part by the Open Fund for Key Laboratory of Image Processing and Pattern Recognition of Jiangxi Province under Grant ET202208305.


\vfill\pagebreak

\bibliographystyle{IEEEbib}
\bibliography{refs}

\begin{thebibliography}{10}

\bibitem{DBLP:journals/ijrr/GeigerLSU13}
Andreas Geiger, Philip Lenz, Christoph Stiller, and Raquel Urtasun,
\newblock ``Vision meets robotics: The {KITTI} dataset,''
\newblock {\em Int. J. Robotics Res.}, vol. 32, no. 11, pp. 1231--1237, 2013.

\bibitem{DBLP:conf/nips/EigenPF14}
David Eigen, Christian Puhrsch, and Rob Fergus,
\newblock ``Depth map prediction from a single image using a multi-scale deep network,''
\newblock in {\em Advances in Neural Information Processing Systems (NeurIPS)}, 2014, pp. 2366--2374.

\bibitem{DBLP:conf/iccv/Yang0DS021}
Guanglei Yang, Hao Tang, Mingli Ding, Nicu Sebe, and Elisa Ricci,
\newblock ``Transformer-based attention networks for continuous pixel-wise prediction,''
\newblock in {\em {IEEE/CVF} International Conference on Computer Vision (ICCV)}, 2021, pp. 16249--16259.

\bibitem{DBLP:conf/cvpr/MiangolehDMPA21}
S.~Mahdi~H. Miangoleh, Sebastian Dille, Long Mai, Sylvain Paris, and Yagiz Aksoy,
\newblock ``Boosting monocular depth estimation models to high-resolution via content-adaptive multi-resolution merging,''
\newblock in {\em {IEEE} Conference on Computer Vision and Pattern Recognition (CVPR)}, 2021, pp. 9685--9694.

\bibitem{DBLP:conf/icra/WofkMYKS19}
Diana Wofk, Fangchang Ma, Tien{-}Ju Yang, Sertac Karaman, and Vivienne Sze,
\newblock ``Fastdepth: Fast monocular depth estimation on embedded systems,''
\newblock in {\em International Conference on Robotics and Automation (ICRA)}, 2019, pp. 6101--6108.

\bibitem{DBLP:journals/tii/TuXLLXHY21}
Xiaohan Tu, Cheng Xu, Siping Liu, Renfa Li, Guoqi Xie, Jing Huang, and Laurence~Tianruo Yang,
\newblock ``Efficient monocular depth estimation for edge devices in internet of things,''
\newblock {\em {IEEE} Trans. Ind. Informatics}, vol. 17, no. 4, pp. 2821--2832, 2021.

\bibitem{DBLP:conf/icra/0006DGNB22}
Michael Rudolph, Youssef Dawoud, Ronja G{\"{u}}ldenring, Lazaros Nalpantidis, and Vasileios Belagiannis,
\newblock ``Lightweight monocular depth estimation through guided decoding,''
\newblock in {\em {IEEE} International Conference on Robotics and Automation (ICRA)}, 2022, pp. 2344--2350.

\bibitem{DBLP:conf/3dim/LainaRBTN16}
Iro Laina, Christian Rupprecht, Vasileios Belagiannis, Federico Tombari, and Nassir Navab,
\newblock ``Deeper depth prediction with fully convolutional residual networks,''
\newblock in {\em International Conference on 3D Vision (3DV)}, 2016, pp. 239--248.

\bibitem{DBLP:conf/cvpr/HeZRS16}
Kaiming He, Xiangyu Zhang, Shaoqing Ren, and Jian Sun,
\newblock ``Deep residual learning for image recognition,''
\newblock in {\em {IEEE} Conference on Computer Vision and Pattern Recognition (CVPR)}, 2016, pp. 770--778.

\bibitem{DBLP:conf/cvpr/GodardAB17}
Cl{\'{e}}ment Godard, Oisin~Mac Aodha, and Gabriel~J. Brostow,
\newblock ``Unsupervised monocular depth estimation with left-right consistency,''
\newblock in {\em {IEEE} Conference on Computer Vision and Pattern Recognition (CVPR)}, 2017, pp. 6602--6611.

\bibitem{DBLP:conf/cvpr/ZhouBSL17}
Tinghui Zhou, Matthew Brown, Noah Snavely, and David~G. Lowe,
\newblock ``Unsupervised learning of depth and ego-motion from video,''
\newblock in {\em {IEEE} Conference on Computer Vision and Pattern Recognition (CVPR)}, 2017, pp. 6612--6619.

\bibitem{DBLP:conf/iccv/GodardAFB19}
Cl{\'{e}}ment Godard, Oisin~Mac Aodha, Michael Firman, and Gabriel~J. Brostow,
\newblock ``Digging into self-supervised monocular depth estimation,''
\newblock in {\em {IEEE/CVF} International Conference on Computer Vision, {ICCV}}, 2019, pp. 3827--3837.

\bibitem{DBLP:conf/cvpr/SandlerHZZC18}
Mark Sandler, Andrew~G. Howard, Menglong Zhu, Andrey Zhmoginov, and Liang{-}Chieh Chen,
\newblock ``Mobilenetv2: Inverted residuals and linear bottlenecks,''
\newblock in {\em {IEEE} Conference on Computer Vision and Pattern Recognition (CVPR)}, 2018, pp. 4510--4520.

\bibitem{DBLP:conf/osdi/ChenMJZYSCWHCGK18}
Tianqi Chen, Thierry Moreau, Ziheng Jiang, Lianmin Zheng, Eddie~Q. Yan, Haichen Shen, Meghan Cowan, Leyuan Wang, Yuwei Hu, Luis Ceze, Carlos Guestrin, and Arvind Krishnamurthy,
\newblock ``{TVM:} an automated end-to-end optimizing compiler for deep learning,''
\newblock in {\em {USENIX} Symposium on Operating Systems Design and Implementation (OSDI)}, 2018, pp. 578--594.

\bibitem{DBLP:journals/corr/abs-2101-06085}
Yuanduo Hong, Huihui Pan, Weichao Sun, and Yisong Jia,
\newblock ``Deep dual-resolution networks for real-time and accurate semantic segmentation of road scenes,''
\newblock {\em arXiv preprint}, arXiv:2101.06085,2021.

\bibitem{DBLP:conf/eccv/WuH18}
Yuxin Wu and Kaiming He,
\newblock ``Group normalization,''
\newblock in {\em European Conference on Computer Vision (ECCV)}, 2018, vol. 11217, pp. 3--19.

\bibitem{DBLP:conf/ijcnn/QinZLZP18}
Zheng Qin, Zhaoning Zhang, Dongsheng Li, Yiming Zhang, and Yuxing Peng,
\newblock ``Diagonalwise refactorization: An efficient training method for depthwise convolutions,''
\newblock in {\em International Joint Conference on Neural Networks (IJCNN)}, 2018, pp. 1--8.

\bibitem{DBLP:conf/eccv/ZhangZPXS18}
Zhenli Zhang, Xiangyu Zhang, Chao Peng, Xiangyang Xue, and Jian Sun,
\newblock ``Exfuse: Enhancing feature fusion for semantic segmentation,''
\newblock in {\em European Conference on Computer Vision (ECCV)}, 2018, vol. 11214, pp. 273--288.

\bibitem{DBLP:conf/iccv/Hu0O19}
Junjie Hu, Yan Zhang, and Takayuki Okatani,
\newblock ``Visualization of convolutional neural networks for monocular depth estimation,''
\newblock in {\em {IEEE/CVF} International Conference on Computer Vision (ICCV)}, 2019, pp. 3868--3877.

\end{thebibliography}

\end{document}